\documentclass[sigconf]{acmart}

\usepackage{booktabs} % For formal tables
\usepackage{amssymb}
\usepackage{amsmath}
\usepackage{multirow}
\usepackage{caption}
\usepackage{subfigure}
\usepackage[ruled,lined,boxed,linesnumbered]{algorithm2e}
\usepackage{color}

\copyrightyear{2019}
\acmYear{2019} 
\setcopyright{iw3c2w3}
\acmConference[WWW '19]{Proceedings of the 2019 World Wide Web Conference}{May 13--17, 2019}{San Francisco, CA, USA}
\acmBooktitle{Proceedings of the 2019 World Wide Web Conference (WWW'19), May 13--17, 2019, San Francisco, CA, USA}
\acmPrice{}
\acmDOI{10.1145/3308558.3313564}
\acmISBN{978-1-4503-6674-8/19/05}
\usepackage{balance}

\newcommand{\ComGAN}{CommunityGAN}

\begin{document}

\title{\ComGAN: Community Detection with Generative Adversarial Nets}

\author{Yuting Jia, Qinqin Zhang, Weinan Zhang, Xinbing Wang}
\affiliation{\institution{Shanghai Jiao Tong University}}
\email{{hnxxjyt,zqq.lh.377,wnzhang,xwang8}@sjtu.edu.cn}

\renewcommand{\shortauthors}{Y. Jia et al.}

\begin{abstract}
Community detection refers to the task of discovering groups of vertices sharing similar properties or functions so as to understand the network data.
With the recent development of deep learning, graph representation learning techniques are also utilized for community detection.
However, the communities can only be inferred by applying clustering algorithms based on learned vertex embeddings.
These general cluster algorithms like K-means and Gaussian Mixture Model cannot output much overlapped communities, which have been proved to be very common in many real-world networks.
In this paper, we propose \ComGAN, a novel community detection framework that jointly solves overlapping community detection and graph representation learning.
First, unlike the embedding of conventional graph representation learning algorithms where the vector entry values have no specific meanings, the embedding of \ComGAN~indicates the membership strength of vertices to communities.
Second, a specifically designed Generative Adversarial Net (GAN) is adopted to optimize such embedding.
Through the minimax competition between the motif-level generator and discriminator, both of them can alternatively and iteratively boost their performance and finally output a better community structure.
Extensive experiments on synthetic data and real-world tasks demonstrate that \ComGAN~achieves substantial community detection performance gains over the state-of-the-art methods.
\end{abstract}




\maketitle

\section{Introduction}

Network is a powerful language to represent relational information among data objects from social, natural and academic domains \cite{yang2014detecting}.
One way to understand network is to identify and analyze groups of vertices which share highly similar properties or functions.
Such groups of vertices can be users from the same organization in social networks \cite{newman2004detecting}, proteins with similar functionality in biochemical networks \cite{krogan2006global}, and papers from the same scientific fields in citation networks \cite{ruan2013efficient}.
The research task of discovering such groups is known as the community detection problem \cite{Zhang2015PNMF}.

Graph representation learning, also known as network embedding, which aims to represent each vertex in a graph as a low-dimensional vector, has been consistently studied in recent years.
The application of deep learning algorithms including Skip-gram \cite{deepwalk,node2vec} and convolutional network \cite{gcn} has improved the efficiency and performance of graph representation learning dramatically.
Moreover, Generative Adversarial Nets (GAN) has also been introduced for learning better graph representation \cite{GraphGAN}.
The learned vertex representation vectors can benefit a wide range of network analysis tasks including link prediction \cite{gao2011temporal,wang2018shine}, recommendation \cite{yu2014personalized,doi:10.1137/1.9781611974973.43}, and node classification \cite{tang2016node}.

However, there still exists many limitations for the application of such embedding in overlapping community detection problems because of the dense overlapping of communities.
Generally, the useful information of vertex embedding vectors is the relevant distance of these vectors, while the specific value in vertex embedding vectors has no meanings.
Thus, given the representation vectors of the vertices, one has to adopt other algorithms like logistic regression to accomplish the real-world application tasks.
To detect communities, a straightforward way is to run some clustering algorithms in the vector space.
However, in some real-world datasets, one vertex may belong to tens of communities simultaneously \cite{DBLP:journals/corr/abs-1110-5813,yang2012community}, while most clustering algorithms cannot handle such dense overlapping.
In recent years, some researchers try to perform the community detection and network embedding simultaneously in a unified framework \cite{M-NMF,ComE} but still fail to solve the dense overlapping problem.

\begin{figure}[]
\includegraphics[width=\linewidth]{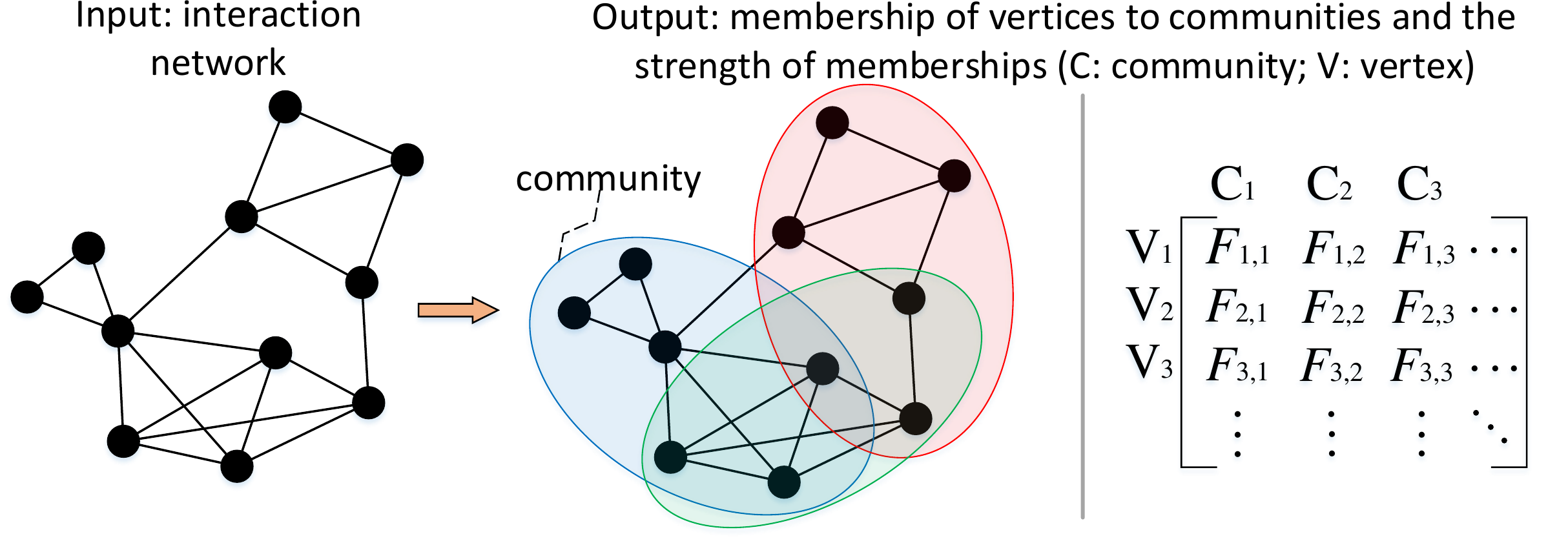}
\caption{The input-output overview of \ComGAN.}
\label{fig:overview}
\end{figure}

In this paper, we propose \textit{\ComGAN}, a novel community detection framework that jointly solves overlapping community detection and graph representation learning.
The input-output overview of \ComGAN~is shown in Figure \ref{fig:overview}.
With \ComGAN, we aim to learn network embeddings like AGM (Affiliation Graph Model) through a specifically designed GAN.
AGM \cite{yang2012community,yang2013overlapping} is a framework which can model densely overlapping community structures.
It assigns each vertex-community pair a nonnegative factor which represents the degree of membership of the vertex to the community.
Thus, the strengths of membership from a vertex to all communities compose the representation vector of it.

Moreover, in recent years, motifs have been proved as essential features in community detection tasks \cite{triangle}.
Thus in this paper, unlike most previous works considering relationship between only two vertices (the relationship between a center vertex and one of the other vertices in a window), we try to generate and discriminate motifs.
Specifically, \ComGAN~trains two models during the learning process:
1) a generator $G(s|v_c)$, which tries to generate the most likely vertex subset $s$ to compose a specified kind of motif;
2) a discriminator $D(s)$, which attempts to distinguish whether the vertex subset $s$ is a real motif or not.
In the proposed \ComGAN, the generator $G$ and the discriminator $D$ act as two players in a minimax game.
Competition in this game drives both of them to improve their capability until the generator is indistinguishable from the true connectivity distribution.

The contributions of our work are threefold:

\begin{itemize}
    \item We combine AGM and GAN in a unified framework, which achieves both the outstanding performance of GAN and the direct vertex-community membership representation in AGM.
    \item We study the motif distribution among ground-truth communities and analyze how they can help improve the quality of detected communities.
    \item We propose a novel implementation for motif generation called \textit{Graph AGM}, which can generate the most likely motifs with graph structure awareness in a computationally efficient way.
\end{itemize}

Empirically, two experiments were conducted on a series of synthetic data, and the results prove: 1) the ability of \ComGAN~to solve dense overlapping problem; 2) the efficacy of motif-level generation and discrimination.
Additionally, to complement these experiments, we evaluate \ComGAN~on two real-world scenarios, i.e., community detection and clique prediction, using five real-world datasets.
The experiment results show that \ComGAN~substantially outperforms the state-of-the-art methods in the field of community detection and graph representation learning.
Specifically, \ComGAN~outperforms baselines by 7.9\% to 21.0\% in terms of F1-Score for community detection.
Additionally, in 3-clique and 4-clique prediction tasks, \ComGAN~improves AUC score to at least 0.990 and 0.956 respectively.
The superiority of \ComGAN~ relies on its joint advantage of particular embedding design, adversarial learning and motif-level optimization.

\section{Related Work}

\textbf{Community Detection.}
Many community detection algorithms have been proposed from different perspectives.
One direction is to design some measure of the quality of a community like modularity, and community structure can be uncovered by optimizing such measures \cite{newman2006modularity,xiang2016local}.
Another direction is to adopt the generative models to describe the generation of the graphs, and the communities can be inferred by fitting graphs to such models \cite{hu2015community,Zhang2015PNMF}.
Moreover, some models focus on the graph adjacency matrix and output the relationship between vertices and communities by adopting matrix factorization algorithms on the graph adjacency matrix \cite{yang2013overlapping,AAAI1817142}.
These models often consider the dense community overlapping problem and detect overlapping communities.
However, the performance of these methods are restricted by performing pair reconstruction with bi-linear models.

\vspace{5pt}\noindent \textbf{Graph representation learning.}
In recent years, several graph representation learning algorithms have been proposed.
For example, DeepWalk \cite{deepwalk} shows the random walk in a graph is similar to the text sequence in natural language. Thus it adopts Skip-gram, a word representation learning model \cite{word2vec}, to learn vertex representation.
Node2vec \cite{node2vec} further extends the idea of DeepWalk by proposing a biased random walk algorithm, which provides more flexibility when generating the sampled vertex sequence.
LINE \cite{LINE} first learns the vertex representation preserving both the first and second order proximities.
Thereafter, GraRep \cite{GraRep} applies different loss functions defined on graphs to capture different $k$-order proximities and the global structural properties of the graph.
Moreover, GAN has also been introduced into graph representation learning.
GraphGAN \cite{GraphGAN} proposes a unified adversarial learning framework, which naturally captures structural information from graphs to learn the graph representation.
ANE \cite{ANE} utilizes GAN as a regularizer for learning stable and robust feature extractor.
However, all of the above algorithms focus on general graph representation learning.
For community detection tasks, we have to adopt other clustering algorithms on vertex embeddings, which cannot handle the dense community overlapping problem.
Compared to above mentioned methods, our \ComGAN~can output the membership of vertices to communities directly.

\vspace{5pt}\noindent \textbf{Unified framework for graph representation learning and community detection.}
In recent years, some unified frameworks for both graph representation learning and community detection have been proposed.
\citet{M-NMF} developed a modularized nonnegative matrix factorization algorithm to preserve both vertex similarity and community structure.
However, because of the complexity of matrix factorization, this model cannot be applied on many real-world graphs containing millions of vertices.
\citet{ComE} designed a framework that jointly solves community embedding, community detection and node embedding together.
However, although the community structures and vertex embeddings are optimized together, the community structures are still inferred through a clustering algorithm (e.g., Gaussian Mixture Model) based on the vertex embedding, which means the dense community overlapping problem cannot be avoided.

\section{Empirical Observation}
In this section, data analysis of the relationship between motifs and communities is presented, which motivates the development of CommunityGAN.
We first describe the network datasets with ground-truth communities and then present our empirical observations.

\subsection{Datasets}

\begin{table}[]
\centering
\caption{Dataset statistics. $V$: number of vertices, $E$: number of edges, $T$: number of 3-clique (triangle), $C$: number of communities, $A$: community memberships per vertex.}
\label{tab:observation-statistics}
\begin{tabular}{l||r|r|r|r|r}
Dataset      & $V$    & $E$    & $T$    & $C$   & $A$   \\
\hline
\hline
LiveJournal  & 4.0M   & 34.9M  & 178M   & 310K  & 3.09  \\
Orkut        & 3.1M   & 120M   & 528M   & 8.5M  & 95.93 \\
Youtube      & 1.1M   & 3.0M   & 3.1M   & 30K   & 0.26  \\
DBLP         & 0.43M  & 1.3M   & 2.2M   & 2.5K  & 2.57  \\
Amazon       & 0.34M  & 0.93M  & 0.67M  & 49K   & 14.83
\end{tabular}
\end{table}

We study the five networks proposed by \citet{yang2012community}, where
LiveJournal, Orkut and Youtube are all online social networks, while
DBLP and Amazon are collaboration network and product network respectively.
Table \ref{tab:observation-statistics} provides the statistics of these networks.

\subsection{Empirical Observations}
Now we present our empirical observations by answering two critical questions.
How do the communities contribute to the generation of motifs?
What is the change in motif generation with communities overlapping?

\begin{table}[]
\centering
\caption{The occurrence probability of cliques for vertices sampled from all vertices or from one community. R: from all vertices. C: from one community.}
\label{tab:observation-probability}
\begin{tabular}{|l|r|r|r|r|r|r|}
\hline
\multirow{2}{*}{Dataset}  & \multicolumn{2}{c|}{2-Clique} & \multicolumn{2}{c|}{3-Clique} & \multicolumn{2}{c|}{4-Clique} \\ \cline{2-7}
             & \multicolumn{1}{c|}{R}    & \multicolumn{1}{c|}{C}  & \multicolumn{1}{c|}{R}    & \multicolumn{1}{c|}{C}  & \multicolumn{1}{c|}{R}   & \multicolumn{1}{c|}{C}   \\ \hline
\hline
LiveJournal  & 4E-6   & 0.80      & 2E-11   & 0.18      & 0        & 0.08       \\ \hline
Orkut        & 2E-5   & 0.82      & 8E-11   & 0.18      & 0        & 0.06       \\ \hline
Youtube      & 4E-6   & 0.73      & 0       & 0.10      & 0        & 0.02       \\ \hline
DBLP         & 1E-5   & 0.52      & 1E-10   & 0.31      & 0        & 0.23       \\ \hline
Amazon       & 6E-6   & 0.53      & 0       & 0.18      & 0        & 0.05       \\ \hline
\end{tabular}
\end{table}

First, we investigate the relationship between motif generation and community structure.
For each time, we randomly select one community.
Then we sample 2/3/4 vertices from this community and judge whether they could compose a motif or not.
We repeat this process for billions of times and get the occurrence probability of motifs in one community.
We compare this value with the motif occurrence  probability in the whole network.
Because, in this paper, we mostly focus on a particular kind of motifs (clique), we only demonstrate the occurrence probability of 2/3/4-vertex cliques.
As shown in Table \ref{tab:observation-probability}, the average occurrence probabilities of cliques for vertices in one community are much higher than that for vertices randomly selected from the whole network.
This observation demonstrates that the occurrence of motifs is strongly correlated to the community structure.

\begin{figure}[tbp]
\centering
\subfigure[3-Clique]{
  \begin{minipage}{0.47\linewidth}
  \centering
  \includegraphics[width=\linewidth]{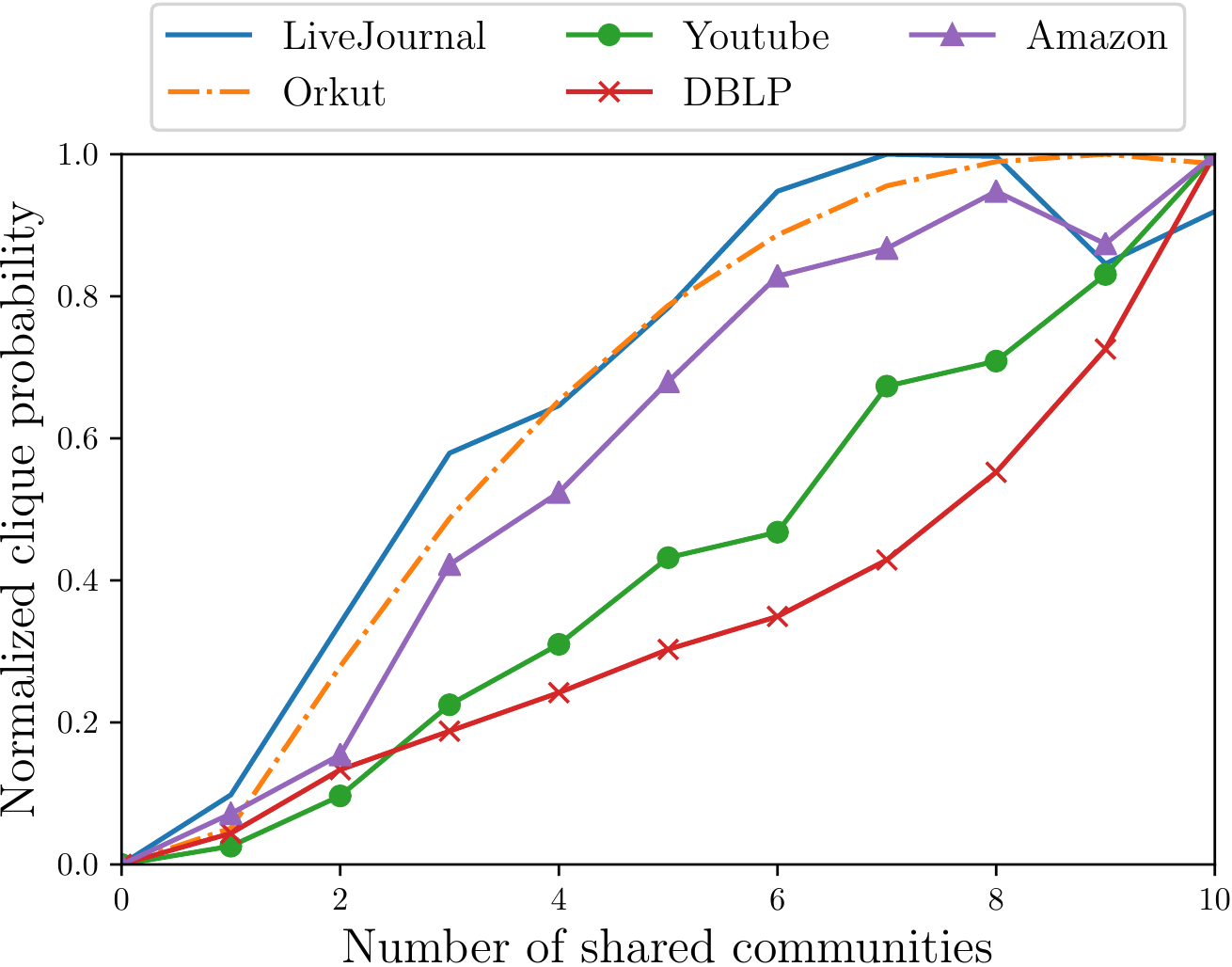}
  \end{minipage}
}
\subfigure[4-Clique]{
  \begin{minipage}{0.47\linewidth}
  \centering
  \includegraphics[width=\linewidth]{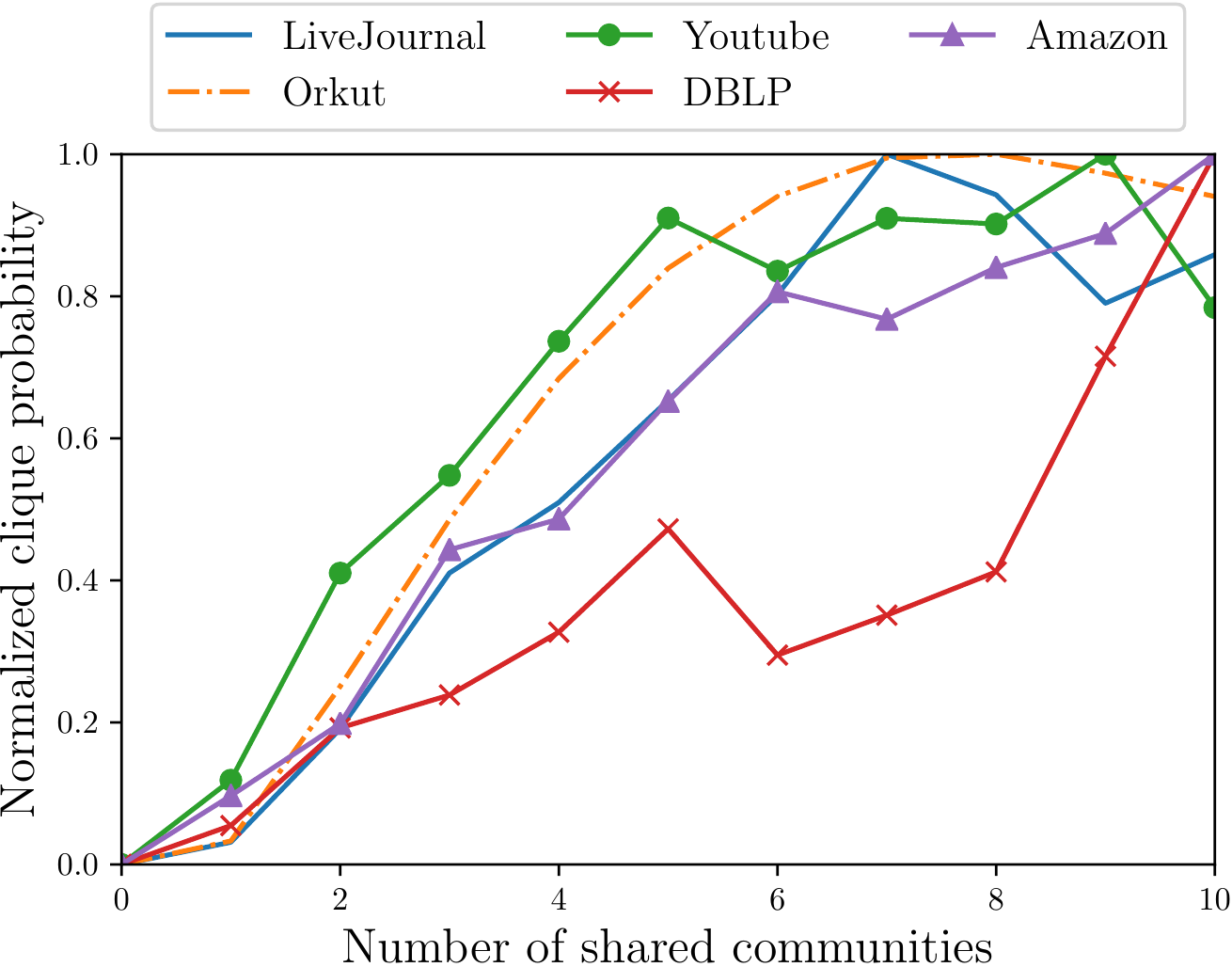}
  \end{minipage}
}
\caption{Normalized probability for vertices to be a clique as a function of number of shared communities. Probabilities are scaled so that maximum value is one.}
\label{fig:clique-probability}
\end{figure}

Next, we study the influence of community overlapping on the generation of motifs.
\citet{yang2013overlapping} have conducted the relationship of 2-clique (edge) probability with community overlapping that the more communities two vertices have in common, the higher probability for them to be 2-clique.
In this paper, we further study the influence of community overlapping on the generation of 3-clique and 4-clique.
As shown in Figure \ref{fig:clique-probability}, similar to 2-clique, the probability curve increases in the overall trend as the number of shared communities increases.
Such observation accords with the base assumption of AGM framework that vertices residing in communities' overlaps are more densely connected to each other than the vertices in a single community.
So we can extend the AGM from edge generation to clique generation, and the details will be explained in \S \ref{sec:graph-agm}.

\section{Methodology}

\textbf{Notation}. 
In this paper, all vectors are column vectors and denoted by lower-case bold letters like $\mathbf{d}_v$ and $\mathbf{g}_v$. 
Calligraphic letters are used to represent sets like $\mathcal{V}$ and $\mathcal{E}$. 
And for simplicity, the corresponding normal letters are used to denote their size like $V=|\mathcal{V}|$. 
The details of notations in this paper are listed in Table \ref{tab:notation}.

\begin{table}[tbp]
  \footnotesize
  \caption{Notations used in this paper}
  \label{tab:notation}
  \begin{tabular}{c|p{4.9cm}}
    \toprule
    Symbol&Description\\
    \midrule
  $\mathcal{G}$       & The studied graph \\
  $\mathcal{V}, \mathcal{E}, \mathcal{M}, \mathcal{C}$       & Set of vertices, edges, motifs, communities in graph $\mathcal{G}$ \\
  $V, E, M, C$        & Number of vertices, edges, motifs, communities in graph $\mathcal{G}$ \\
  $M(v_c)$            & Set of motifs covering $v_c$ in graph $\mathcal{G}$  \\
  $N(v_c)$            & Set of neighbors of vertex $v_c$ in graph $\mathcal{G}$  \\
  $s$                 & The subset of vertices from $\mathcal{V}$ \\
  $p_{true}(m|v_c)$   & The preference distribution of motifs covering $v_c$ over all other motifs in $\mathcal{M}$ \\
  $G(s|v_c;\mathbf{\theta}_G)$ & Generator trying to generate vertex subsets from $\mathcal{V}$ covering $v_c$ most likely to be real motifs \\
  $D(s, \mathbf{\theta}_D)$    & Discriminator aiming to estimate the probability that a vertex subset $s$ is a real motif  \\
  $\mathbf{d}_v, \mathbf{g}_v \in \mathbb{R}_+^C$ & Nonnegative $C$-dimensional representation vector of vertex $v$ in discriminator $D$ and generator $G$, respectively \\
  $\mathbf{\theta}_D, \mathbf{\theta}_G \in \mathbb{R}_+^{V \times C}$ & The union of all $\mathbf{d}_v$ and $\mathbf{g}_v$, respectively \\
  $F_{uc} \in \mathbb{R}_+$            & The nonnegative strength for vertex $u$ to be affiliated to community $c$ used in AGM \\
  $p_c(v_1, v_2, \ldots , v_m)$ & The probability of $m$ vertices $v_1$ to $v_m$ to be a clique through of community $c$ \\
  $p(v_1, v_2, \ldots , v_m)$ & The probability of $m$ vertices $v_1$ to $v_m$ to be a clique through of any one community \\
  $\odot(F_{v_1},F_{v_2},\ldots ,F_{v_m})$ & Sum of the entrywise product from $F_{v_1}$ to $F_{v_m}$, i.e., $\sum_{c=1}^{C}\prod_{i=1}^{m}F_{v_ic}$ \\
  $G_v(v_{s_m}|v_{s_1},\ldots ,v_{s_{m-1}})$ & The vertex generator based on selected vertices $v_{s_1},\ldots ,v_{s_{m-1}}$ in vertex subset generation process\\
  $p_{v_v}(v_i|v_c)$ & The relevance probability of $v_i$ given $v_c$ with root $v_v$ in random walk process \\
  \bottomrule
\end{tabular}
\end{table}

In this section, we introduce the framework of \ComGAN~and discuss the implementation and optimization of the generator and discriminator in \ComGAN.

\subsection{\ComGAN~Framework}

Let $\mathcal{G} = (\mathcal{V}, \mathcal{E}, \mathcal{M})$ represent the studied graph, where $\mathcal{V}=\{v_1,\ldots ,v_V\}$ is a set of vertices, $\mathcal{E}=\{e_{ij}\}_{i,j=1}^V$ is the set of edges among the vertices, and $\mathcal{M}$ is the set of motifs in graph $\mathcal{G}$.
In this paper, notably, we only focus on a particular kind of motifs: cliques.
Moreover, the $D$ and $G$ are also designed to discriminate and generate cliques.
For a given vertex $v_c$, we define $M(v_c)$ as the set of motifs in the graph covering $v_c$, whose size is generally much smaller than the total number of vertices $V$.
Moreover, we define conditional probability $p_{true}(m|v_c)$ as the preference distribution of motifs covering $v_c$ over all other motifs in $\mathcal{M}$.
Thus $M(v_c)$ can be seen as a set of observed motifs drawn from $p_{true}(m|v_c)$.
Given the graph $\mathcal{G}$, we aim to learn the following two models:
\textbf{Generator} $G(s|v_c;\mathbf{\theta}_G)$, which tries to approximate $p_{true}(m|v_c)$, and generate (or select) subsets of vertices from $\mathcal{V}$ covering $v_c$ most likely to be real motifs; and
\textbf{Discriminator} $D(s, \mathbf{\theta}_D)$, which aims to estimate the probability that a vertex subset $s$ is a real motif, i.e., comes from $\mathcal{M}$.

The generator $G$ and discriminator $D$ are combined by playing a minimax game:
generator $G$ would try to perfectly approximate $p_{true}(m|v_c)$ and generate the most likely vertex subsets similar to real motifs covering $v_c$ to fool the discriminator,
while discriminator $D$ would try to distinguish between ground-truth motifs from $p_{true}(m|v_c)$ and the ones generated by $G$.
Formally, $G$ and $D$ act as two opponents in the following two-player minimax game with the joint value function $V(G, D)$:
\begin{equation}
\label{eq:loss-func}
\small
\begin{aligned}
  \min_{\mathbf{\theta}_G}\max_{\mathbf{\theta}_D}&V(G, D)=\sum_{c=1}^V \Big(\mathbb{E}_{m \sim p_{true}(\cdot|v_c)}[\log D(m;\mathbf{\theta}_D)]\\
  &+\mathbb{E}_{s \sim G(s|v_c;\mathbf{\theta}_G)}[\log (1 - D(s;\mathbf{\theta}_D))] \Big).
\end{aligned}
\end{equation}

\begin{figure}[tbp]
\includegraphics[width=\linewidth]{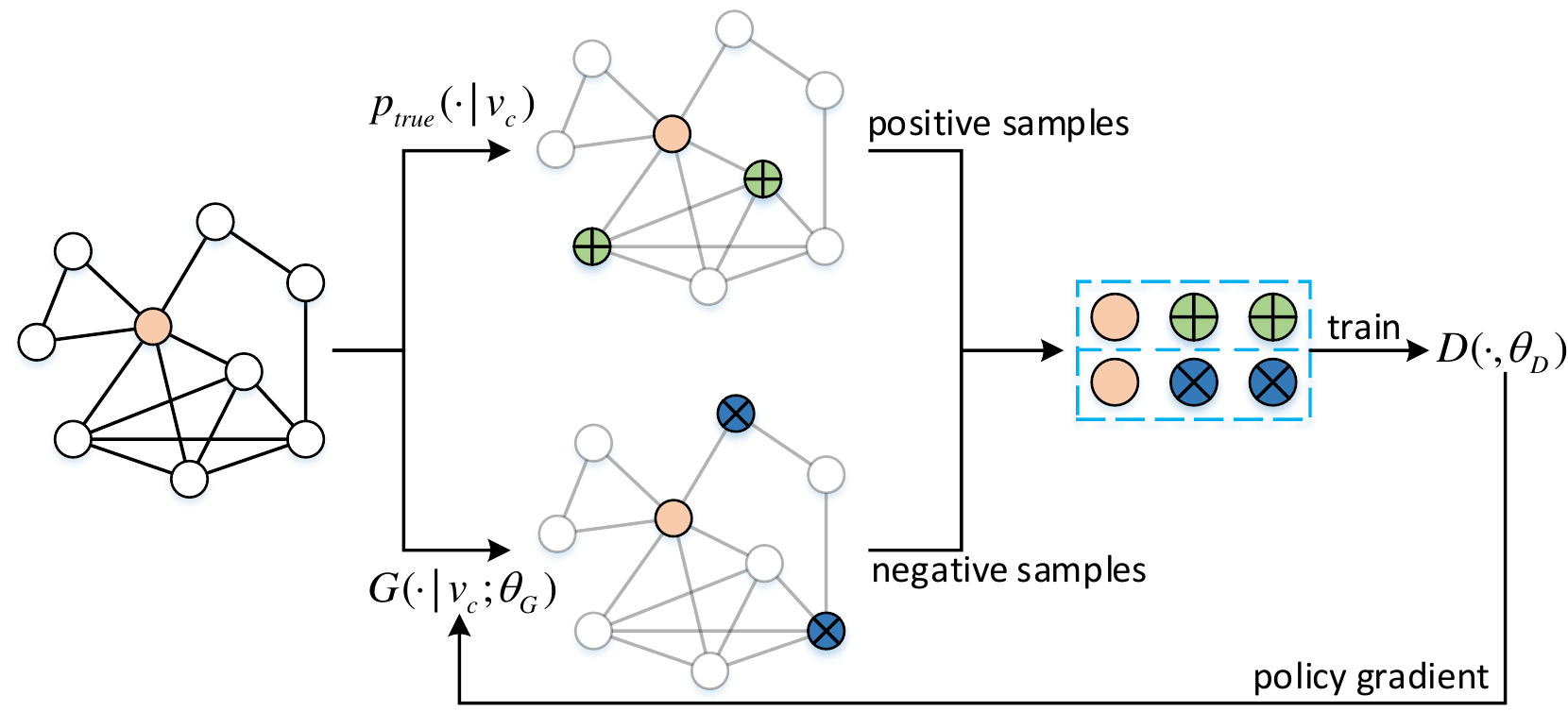}
\caption{Framework of \ComGAN}
\label{fig:framework}
\end{figure}

Based on Eq. (\ref{eq:loss-func}), the optimal $G$ and $D$ can be learned by alternatively maximizing ($D$) and minimizing ($G$) the value function $V(G, D)$.
The \ComGAN~framework is illustrated in Figure \ref{fig:framework}.
Discriminator $D$ is trained with positive samples from $p_{true}(\cdot|v_c)$ and negative samples from $G(\cdot|v_c;\mathbf{\theta}_G)$, and generator $G$ is updated with policy gradient technique \cite{sutton2000policy} under the guidance of $D$ (detailedly described later in this section).
Competition between $G$ and $D$ drives both of them to improve until $G$ is indistinguishable from the true motif distribution $p_{true}$.

\subsection{\ComGAN~Optimization}
Given positive samples from true motif distribution $p_{true}$ and negative samples from the generator $G$, the discriminator aims to maximize the log-probability of correctly classifying these samples, which could be solved by gradient ascent if $D$ is differentiable for $\mathbf{\theta}_D$, i.e.,
\begin{equation}
  \small
  \label{eq:p-V-p-D}
  \begin{aligned}
    \nabla_{\mathbf{\theta}_D}V(G,D) = & \sum_{c=1}^V \Big(\mathbb{E}_{m \sim p_{true}(\cdot|v_c)}[\nabla_{\mathbf{\theta}_D} \log D(m;\mathbf{\theta}_D)] \\
    &+\mathbb{E}_{s \sim G(s|v_c;\mathbf{\theta}_G)}[\nabla_{\mathbf{\theta}_D} \log (1 - D(s;\mathbf{\theta}_D))] \Big).
  \end{aligned}
\end{equation}

In contrast to the discriminator, the objective of the generator is to minimize the log-probability that the discriminator correctly distinguishes samples from $G$.
Because the sampling of $s$ is discrete, we follow \cite{schulman2015gradient,yu2017seqgan,GraphGAN} to compute the gradient of $V(G,D)$ with respect to $\mathbf{\theta}_G$ by policy gradient:
\begin{equation}
  \small
  \label{eq:p-V-p-G}
  \begin{aligned}
     \nabla_{\mathbf{\theta}_G} V(G,D) &=\nabla_{\mathbf{\theta}_G} \sum_{c=1}^V{\mathbb{E}_{s \sim G(\cdot|v_c)}[\log(1 - D(s))]} \\
    &=\sum_{c=1}^V{\mathbb{E}_{s \sim G(\cdot|v_c)}[\nabla_{\mathbf{\theta}_G} \log G(s|v_c)\log(1 - D(s))]}.
  \end{aligned}
\end{equation}
\subsection{A Naive Implementation of $D$ and $G$}

A naive implementation of the discriminator and generator is based on sigmoid and softmax functions, respectively.

For the discriminator $D$, intuitively we can define it as the multiplication of the sigmoid function of the inner product of every two vertices in the input vertex subset $s$:

\begin{equation}
\label{eq:d(s)}
\small
  D(s)  = \prod_{(u,v) \in s, u \neq v}{\sigma(\mathbf{d}_u^\top  \cdot \mathbf{d}_v)},
\end{equation}
where $\mathbf{d}_u, \mathbf{d}_v \in \mathbb{R}^k$ are the $k$-dimensional representation vectors for discriminator $D$ of vertices $u$ and $v$ respectively, and $\mathbf{\theta}_D$ is the union of all $\mathbf{d}_v$'s.

For the implementation of $G$,
to generate a vertex subset $s$ covering vertex $v_c$, we can regard the subset as a sequence of vertices $(v_{s_1},v_{s_2},\ldots ,v_{s_m})$ where $v_{s_1}=v_c$.
Then the generator $G$ can be defined as follows:
\begin{equation}
  \small
  \label{eq:g}
  \begin{aligned}
    &G(s|v_c)  \\
    =&G_v(v_{s_2}|v_{s_1})G_v(v_{s_3}|v_{s_1},v_{s_2})\cdots G_v(v_{s_m}|v_{s_1},\ldots ,v_{s_{m-1}}).
  \end{aligned}
\end{equation}

Notably, in Eq. (\ref{eq:g}), the generation of $v_{s_m}$ is based on vertices from $v_{s_1}$ to $v_{s_{m-1}}$, not only on $v_{s_{m-1}}$.
Because if we generate $v_{s_{m}}$ based only on $v_{s_{m-1}}$, it will be very likely that vertex $v_{s_{m}}$ and the other vertices belong to different communities.
For example, vertices from $v_{s_{1}}$ to $v_{s_{m-1}}$ are students in one university, while vertex $v_{s_{m}}$ is the parent of $v_{s_{m-1}}$.
Simply, we can know the probability of the vertex subset $s$ being a clique will be very low.
Thus, we generate $v_{s_m}$ based on all the vertices from $v_{s_1}$ to $v_{s_{m-1}}$.

For the implementation of the vertex generator $G_v$, straightforwardly, we can define it as a softmax function over all other vertices, i.e.,
\begin{equation}
\label{eq:gv}
\small
  G_v(v_{s_m}|v_{s_1},\ldots ,v_{s_{m-1}}) = \frac{{\exp(\sum_{i=1}^{m-1}\mathbf{g}_{v_{s_m}}^\top \mathbf{g}_{v_{s_i}})}}{\sum_{v \notin (v_{s_1},\ldots ,v_{s_{m-1}})}{{\exp(\sum_{i=1}^{m-1}\mathbf{g}_v^\top \mathbf{g}_{v_{s_i}})}}},
\end{equation}
where $\mathbf{g}_v \in \mathbb{R}^k$ is the $k$-dimensional representation vectors for generator $G_v$ of vertex $v$, and $\mathbf{\theta}_G$ is the union of all $\mathbf{g}_v$'s.

\subsection{Graph AGM}
\label{sec:graph-agm}

Sigmoid and softmax function provide concise and intuitive definitions for the motif discrimination in $D_s$ and vertex generation in $G_v$, but they have three limitations in community detection task:
1) To detect community, after learning the vertex representation vectors based on Eq. (\ref{eq:d(s)}) and (\ref{eq:gv}), we still need to adopt some clustering algorithms to detect the communities.
According to \cite{DBLP:journals/corr/abs-1110-5813}, the overlap is indeed a significant feature of many real-world social networks.
Moreover, in \cite{yang2012community}, the authors showed that in some real-world datasets, one vertex might belong to tens of communities simultaneously.
However, general clustering algorithms cannot handle such dense overlapping.
2) The calculation of softmax in Eq. (\ref{eq:gv}) involves all vertices in the graph, which is computationally inefficient.
3) The graph structure encodes rich information of proximity among vertices, but the softmax in Eq. (\ref{eq:gv}) completely ignores it.

To address these problems, in \ComGAN~we propose a novel implementation for the discriminator and generator, which is called Graph AGM.

\begin{figure}[]
\includegraphics[width=\linewidth]{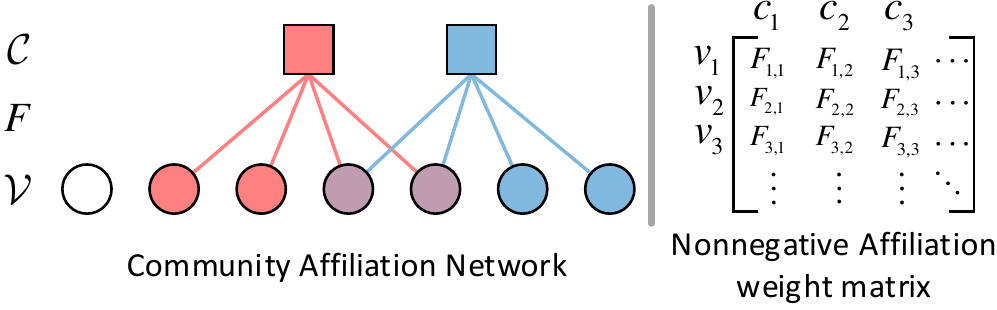}
\caption{AGM framework. $\mathcal{V}$: vertices; $\mathcal{C}$: communities; $F$: affiliation of vertices to communities.}
\label{fig:AGM}
\end{figure}

\begin{figure*}[t]
\includegraphics[width=\linewidth]{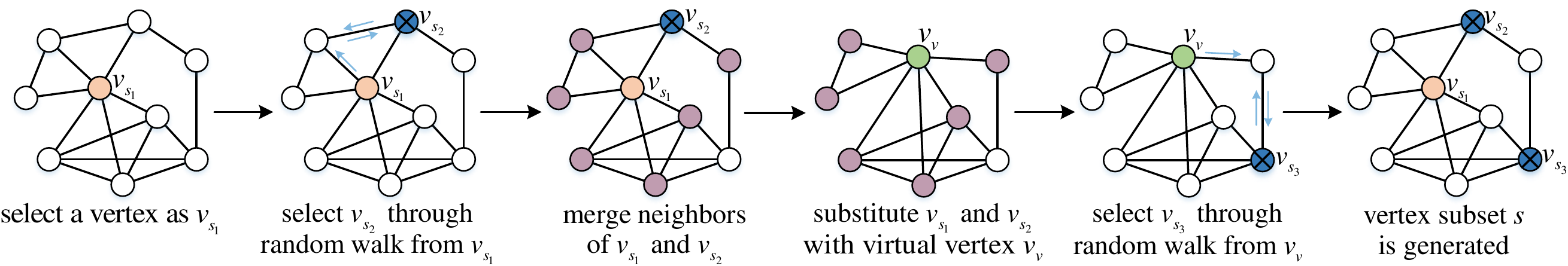}
\caption{The generation process for vertex subset $s$. Blue arrows indicate the path of random walk. At the two blue vertices, because the vertex generator decides to visit the previous vertex, the random walk finishes and the blue vertices are selected.}
\label{fig:random_walk}
\end{figure*}

AGM (Affiliation Graph Model) \cite{yang2012community,yang2013overlapping} is based on the idea that communities arise due to shared group affiliation, and views the whole network as a result generated by a community-affiliation graph model.
The framework of AGM is illustrated in Figure \ref{fig:AGM}, which can be either seen as a bipartite network between vertices and communities, or written as a nonnegative affiliation weight matrix.
In AGM, each vertex could be affiliated to zero, one or more communities.
If vertex $u$ is affiliated to community $c$, there will be a nonnegative strength $F_{uc}$ of this affiliation.
For any community $c \in \mathcal{C}$, it connects its member vertices $u, v$ with probability $1-\exp(-F_{uc} \cdot F_{vc})$.
Moreover, each community $c$ creates edges independently.
If the pair of vertices $u, v$ are connected multiple times through different communities, the probability is $1-\exp(-\sum_cF_{uc} \cdot F_{vc})$.
So that the probability that vertices $u, v$ are connected (through any possible communities) is $p(u,v) = 1-\exp(-F_{u}^\top  \cdot F_{v})$, where $F_u$ and $F_v$ are the nonnegative $C$-dimensional affiliation vectors for vertices $u$ and $v$ respectively.

We extend AGM from edge generation to motif generation.
For any $m$ vertices $v_1$ to $v_m$, we assume that the probability of them to be a clique through of community $c$ is defined as $p_c(v_1, v_2,\ldots , v_m) = 1 - \exp(-\prod_{i=1}^{m} F_{v_{i} c})$.
Then the probability that these $m$ vertices compose a clique through any possible communities can be calculated via
\begin{equation}
\label{eq:p-agm-clique}
\small
\begin{aligned}
  p(v_1, v_2,\ldots , v_m)  &= 1 - \prod_{c}{(1 - p_c(v_1, v_2,\ldots , v_m))}\\
              &= 1 - {\exp(-\odot(F_{v_1},F_{v_2},\ldots ,F_{v_m}))},\\
\end{aligned}
\end{equation}
where $\odot(F_{v_1},F_{v_2},\ldots ,F_{v_m})$ means the sum of the entrywise product from $F_{v_1}$ to $F_{v_m}$, i.e., $\sum_{c=1}^{C}\prod_{i=1}^{m}F_{v_ic}$.

Then the discriminator, which was defined as the product of sigmoid in a straightforward way, can be redefined as
\begin{equation}
\label{eq:d(s)-agm}
\small
  D(s)  = 1 - {\exp(-\odot(\mathbf{d}_{v_1},\mathbf{d}_{v_2},\ldots ,\mathbf{d}_{v_m}))},
\end{equation}
where $\mathbf{d}_v \in \mathbb{R}^C$ is the nonnegative $C$-dimensional representation vectors of vertex $v$ for discriminator $D$, and $\mathbf{\theta}_D$ is the union of all $\mathbf{d}_v$'s.

Moreover, the generator $G_v$ can be redefined as the softmax function over all other possible vertices to compose a clique with $m-1$ chosen vertices:
\begin{equation}
  \small
  \label{eq:gv-agm}
  \begin{aligned}
    &G_v(v_{s_m}|v_{s_1},\ldots ,v_{s_{m-1}})  \\
    &= \frac{1 - \exp(-\odot(\mathbf{g}_{v_{s_1}},\ldots ,\mathbf{g}_{v_{s_m}}))}{\sum_{v \notin (v_{s_1},\ldots ,v_{s_{m-1}})}{1 - \exp(-\odot(\mathbf{g}_{v_{s_1}},\ldots ,\mathbf{g}_{v_{s_{m-1}}},\mathbf{g}_{v}))}},
  \end{aligned}
\end{equation}
where $\mathbf{g}_v \in \mathbb{R}^C$ is the nonnegative $C$-dimensional representation vectors of vertex $v$ for generator $G_v$, and $\mathbf{\theta}_G$ is the union of all $\mathbf{g}_v$'s.
With this setting, the learned vertex representation vectors $\mathbf{g}_v$ will represent the affiliation weight between vertex $v$ and communities, which means we need no additional clustering algorithms to find the communities and the first aforementioned limitation is omitted.

To further overcome the other two limitations, inspired by \cite{GraphGAN}, we design graph AGM as follows.
To calculate $G_v(v_{s_m}|v_{s_1},\ldots,v_{s_{m-1}})$, we first assume there is a virtual vertex $v_v$ which is connected to all the vertices in the union of neighbors of vertices from $v_{s_1}$ to $v_{s_{m-1}}$, i.e., $N(v_v)=N(v_{s_1})\cup\cdots \cup N(v_{s_{m-1}})$ where $N(v)$ represents the set of neighbors of vertex $v$.
Moreover, we assign $\mathbf{g}_{v_v} = \mathbf{g}_{v_{s_1}} \circ\cdots\circ \mathbf{g}_{v_{s_{m-1}}}$, i.e., the representation vector of $v_v$ is the entrywise product of the representation vectors of vertices from $v_{s_1}$ to $v_{s_{m-1}}$.
For simplicity, we substitute $G_v(v_{s_m}|v_{s_1},\ldots ,v_{s_{m-1}})$ with $G_v(v_{s_m}|v_v)$.
Then we adopt random walk starting from the virtual vertex $v_v$ based on a well designed probability distribution (Eq. (\ref{eq:p-v-v})).
During the process of random walk, if the currently visited vertex is $v$ and generator G decides to visit $v$'s previous vertex, then $v$ is chosen as the generated vertex and the random walk process stops.
The whole process for generating a vertex subset with the size of 3 is illustrated in Figure \ref{fig:random_walk}.

\begin{algorithm}[t]
\small
\caption{\ComGAN~framework}
\label{alg:framework}
\KwIn{number of communities $c$, size of discriminating samples $m$, size of generating samples $n$.}
\KwOut{generator $G(s|v_c;\mathbf{\theta}_G)$, discriminator $D(s, \mathbf{\theta}_D)$.}

Initialize and pre-train $G(s|v_c;\mathbf{\theta}_G)$ and $D(s, \mathbf{\theta}_D)$\;
\While{\ComGAN~not converge}{
  \For{G-steps}{
    Generate $n$ vertex subsets from $G(s|v_c;\mathbf{\theta}_G)$ for each vertex $v_c$ based on the random walk process\;
    Update $\mathbf{\theta}_G$ according to Eq. (\ref{eq:p-V-p-G}), (\ref{eq:g}), (\ref{eq:p-v-v}) and (\ref{eq:G_v_final})\;
  }
  \For{D-steps}{
    Sample $m$ positive vertex subsets from $p_{true}$ and $s$ negative vertex subsets from $G(s|v_c;\mathbf{\theta}_G)$ for each vertex $v_c$\;
    Update $\mathbf{\theta}_D$ according to Eq. (\ref{eq:p-V-p-D}) and (\ref{eq:d(s)-agm})\;
  }
}
\Return $G(s|v_c;\mathbf{\theta}_G)$ and $D(s, \mathbf{\theta}_D)$\;
\end{algorithm}

Moreover, in the random walk process, we wish the walk path is always relevant to the root vertex $v_v$ for maximizing the probability of the generated vertex subset to be a motif.
So that for a given vertex $v_c$ and one of its neighbors $v_i \in N(v_c)$, we define the relevance probability of $v_i$ given $v_c$ as
\begin{equation}
\label{eq:p-v-v}
\small
  p_{v_v}(v_i|v_c) = \frac{1 - \exp(-\odot(\mathbf{g}_{v_i},\mathbf{g}_{v_c},\mathbf{g}_{v_v}))}{\sum_{v_j \in N(v_c)}{1 - \exp(-\odot(\mathbf{g}_{v_j},\mathbf{g}_{v_c},\mathbf{g}_{v_v}))}},
\end{equation}
which is actually a softmax function over $N(v_c)$ for composing clique with vertices $v_v$ and $v_c$.

If we denote the path of random walk as $P_r=(v_{r_1}, v_{r_2},\ldots , v_{r_n})$ where $v_{r_1}=v_v$, the probability for selecting this path will be $p_{v_v}(v_{r_{n-1}}|v_{r_n}) \cdot \prod_{i=1}^{n-1}p_{v_v}(v_{r_{i+1}}|v_{r_i})$.
In the policy gradient, we regard the selection of this path as an action and the target is to maximize its reward from $D$.
Thus, although there may be multiple paths between $v_{r_1}$ and $v_{r_n}$, if we have selected the path $P_r$, we will optimize the policy gradient on it and neglect other paths.
In other words, if we select the path $P_r$, we assign $G_v(v_{s_m}|v_v)$ as follows:
\begin{equation}
\label{eq:G_v_final}
\small
  G_v(v_{s_m}|v_v) = p_{v_v}(v_{r_{n-1}}|v_{r_n})\cdot \prod_{i=1}^{n-1}p_{v_v}(v_{r_{i+1}}|v_{r_i}).
\end{equation}

Finally, the overall solution of \ComGAN~is summarized in Algorithm \ref{alg:framework}.

\subsection{Other Issues}

\vspace{5pt}\noindent\textbf{Model Initialization.}
We have two methods to initialize the generator $G$ and the discriminator $D$.
The first is that we can deploy AGM model on the graph to learn a community affiliation vector $F_i$ for each vertex $v_i$, and then we can set $g_{v_i}=d_{v_i}=F_i$ directly.
The second is that we can use locally minimal neighborhoods \cite{Neighborhoods-2012} to initialize $\mathbf{\theta}_G$ and $\mathbf{\theta}_D$.
We can regard each vertex $v_i$ along with its neighbors $N(v_i)$, denoted as $C(v_i)$, as a community.
Community $C_(v_i)$ is called locally minimal if $C_(v_i)$ has lower conductance than all the $C_(v_j)$ for vertices $v_j$ who are connected to vertex $v_i$.
\citet{Neighborhoods-2012} have empirically showed that the locally minimal neighborhoods are good seed sets for community detection algorithms.
For a node $v_i$ who belongs to a locally minimal neighborhood $c$, we initialize $F_{v_i}c = 1$, otherwise $F_{v_i}c = 0$.

The second method can save the training time of AGM on the graph.
However, we find that the performance of such initialization is a little lower than the first initialization.
To achieve best performance, in this paper, we choose to adopt the first initialization method.
Moreover, in the efficiency analysis experiment (refer to \S \ref{efficiency-analysis}), the training time of \ComGAN~includes the time of the pre-training process.

\vspace{5pt}\noindent\textbf{Determining community membership.} 
After learning parameters $\mathbf{\theta}_G$ and $\mathbf{\theta}_D$, we need to determine the ``hard'' community membership of each node.
We achieve this by thresholding $\mathbf{\theta}_G$ and $\mathbf{\theta}_D$ with a threshold $\delta$.
The basic intuition is that if two nodes belong to the same community $c$, then the probability of having a link between them through community $c$ should be larger than the background edge probability $\epsilon = 2E/V(V-1)$.
Following this idea, we can obtain the threshold $\delta$ as below:
\begin{equation}
\label{threshold}
\small
  1 - \exp(-\delta^2) = \epsilon \Rightarrow \delta = \sqrt{-\log(1-\epsilon)}
\end{equation} 

With $\delta$ obtained, we consider vertex $v_i$ belonging to community $c$ if ${\mathbf{g}_{v_i}}_c \ge \delta$ or ${\mathbf{d}_{v_i}}_c \ge \delta$ for generator $G$ and discriminator $D$ respectively. 

\vspace{5pt}\noindent\textbf{Choosing the number of communities.}
We follow the method proposed in \cite{airoldi2009mixed} to choose the number of communities $C$.
Specifically, we reserve 20\% of links for validation and learn the model parameters with the remaining 80\% of links for different $C$.
After that, we use the learned parameters to predict the links in validation set and select the $C$ with the maximum prediction score as the number of communities.

\section{Synthetic Data Experiments}

In this section, in order to evaluate the effectiveness of \ComGAN, we conduct a series of experiments based on synthetic datasets.
Specifically, we conduct two experiments to prove: 1) the ability of \ComGAN~to solve the dense overlapping problem; 2) the efficacy of motif-level generation and discrimination.
To simulate the real-world graphs with ground-truth overlapping communities, we adopt CKB Graph Generator - a method that can generate large random social graphs with realistic overlapping community structure \cite{ckb-benchmark} - to generate such synthetic graphs.
The code of \ComGAN~(including pre-training model) and demo datasets are available online\footnote{https://github.com/SamJia/CommunityGAN}.

\subsection{Evaluation Metric}
\label{subsection:evaluation metric}
The availability of ground-truth communities allows us to quantitatively evaluate the performance of community detection algorithms.
Given a graph $\mathcal{G} = (\mathcal{V}, \mathcal{E}, \mathcal{M})$, we denote the set of ground-truth communities as $\mathcal{C}^*$ and the set of detected communities as $\hat{\mathcal{C}}$, where each ground-truth community $c^*_i \in \mathcal{C}^*$ and each detected community $\hat{c}_i \in \hat{\mathcal{C}}$ are both defined by a set of their member nodes.
We consider the average F1-Score as the metric to measure the performance of the detected communities $\hat{\mathcal{C}}$.
To compute the F1-Score between detected communities and ground-truth communities, we need to determine which $c^*_i \in \mathcal{C}^*$ corresponds to which $\hat{c}_j \in \hat{\mathcal{C}}$.
We define final F1-Score as the average of the F1-score of the best-matching ground-truth community to each detected community, and the F1-score of the best-matching detected community to each ground-truth community:
\begin{equation}
\small
\begin{aligned}
  F1(\mathcal{C}^*, \hat{\mathcal{C}}) = \frac{1}{2}(\frac{1}{C^*}\sum_{c^*_i \in \mathcal{C}^*}\max_{\hat{c}_j \in \hat{\mathcal{C}}} F1(c^*_i, \hat{c}_j) + \frac{1}{\hat{C}}\sum_{\hat{c}_i \in \hat{\mathcal{C}}}\max_{c^*_j \in \mathcal{C}^*} F1(c^*_j, \hat{c}_i))
\end{aligned}
\end{equation}
The higher F1-Score means that the detected communities are more accurate and have better qualities.

\begin{table}[tbp]
\centering
\caption{Synthetic Graphs statistics. $V$: number of vertices, $E$: number of edges, $C$: number of communities, $A$: average number of community memberships per vertex, $P$: percentage of communities that have overlapping with others}
\label{tab:synthetic-dataset-statistics}
\begin{tabular}{l||r|r|r|r|r}
Graph   & $V$  & $E$   & $C$ & $A$  & $P$   \\
\hline
\hline
ckb-290 & 1078 & 29491 & 226 & 2.76 & 100\% \\
ckb-280 & 1044 & 30055 & 252 & 3.03 & 100\% \\
ckb-276 & 1023 & 38775 & 339 & 3.74 & 100\% \\
ckb-273 & 1018 & 28025 & 372 & 3.94 & 100\% \\
ckb-270 & 1007 & 30740 & 371 & 4.20 & 100\% \\
ckb-267 & 1123 & 65813 & 381 & 4.57 & 100\% 
\end{tabular}
\end{table}

\subsection{Datasets}
To prove \ComGAN's ability to solve the dense overlapping problem, we utilize the CKB Graph Generator to generate synthetic graphs with ground-truth communities of different overlapping levels.
All the parameters of CKB are set as default except two: `power law exponent of user-community membership distribution ($\beta_1$)' and `power law exponent of community size distribution ($\beta_2$)' \cite{ckb-benchmark}.
By setting $\beta_1 = \beta_2 = {2.9, 2.8, 2.76, 2.73, 2.7, 2.67}$, we can get a series of graphs with different levels of community overlapping.
The detailed statistics of these graphs are shown in Table \ref{tab:synthetic-dataset-statistics}.
We can see that the percentages of overlapped communities (denoted by $P$) are all $100\%$, indicating that all the communities overlap more or less with others.
Besides, with the decrease of $\beta_1$ and $\beta_2$, the average number of community memberships per vertex (denoted by $A$) increases, which means the density of community overlapping is becoming higher and higher.

\subsection{Comparative Methods}

We compare \ComGAN~with $7$ baseline methods, of which $3$ are traditional overlapping community detection methods (MMSB, CPM and AGM), $3$ are recent graph representation learning methods (node2vec, LINE and GraphGAN), and $1$ combines graph representation learning as well as community detection (ComE).

\begin{description}
\item[MMSB] \cite{airoldi2009mixed} is one of the representatives of overlapping community detection methods based on dense subgraph extraction.
\item[CPM] \cite{CPM} builds up the communities from k-cliques and allows overlapping between the communities in a natural way.
\item[AGM] \cite{yang2013overlapping} is a framework which can model densely overlapping community structures.
\item[Node2vec] \cite{node2vec} adopts biased random walk and Skip-Gram to learn vertex embeddings.
\item[LINE] \cite{LINE} preserves the first-order and second-order proximity among vertices in the graph.
\item[GraphGAN] \cite{GraphGAN} unifies generative and discriminative graph representation learning methodologies via adversarial training in a minimax game.
\item[ComE] \cite{ComE} jointly solves the graph representation learning and community detection problem.
\end{description}

\subsection{Experiment Setup}
\label{subsec:experiment-setup}
For all the experiments, we perform stochastic gradient descent to update parameters with learning rate $0.001$.
In each iteration, $m$ and $n$ are both 5, i.e., 5 positive and 5 negative vertex subsets will be sampled or generated for each vertex (refer to Algorithm \ref{alg:framework}), and then we update $G$ and $D$ on these vertex subsets for 3 times.
The final learned communities and vertex representations are $\mathbf{g}_v$'s.
Hyperparameter settings for all baselines are as default.

For \ComGAN, CPM, AGM, MMSB and ComE, which directly output the communities for vertices, we fit the network into them to detect communities directly.
For node2vec, LINE and GraphGAN, which can only output the representation for vertices, we fit the network into them and apply K-means to the learned embeddings to get the communities.
Because of the randomness of K-means, we repeat 5 times and report the average results.
Moreover, the number of ground-truth communities is set as the input parameters to the community detection models and K-means.

\subsection{Ability of \ComGAN~to Solve Dense Overlapping Problem}

\begin{figure}[tbp]
\includegraphics[width=\linewidth]{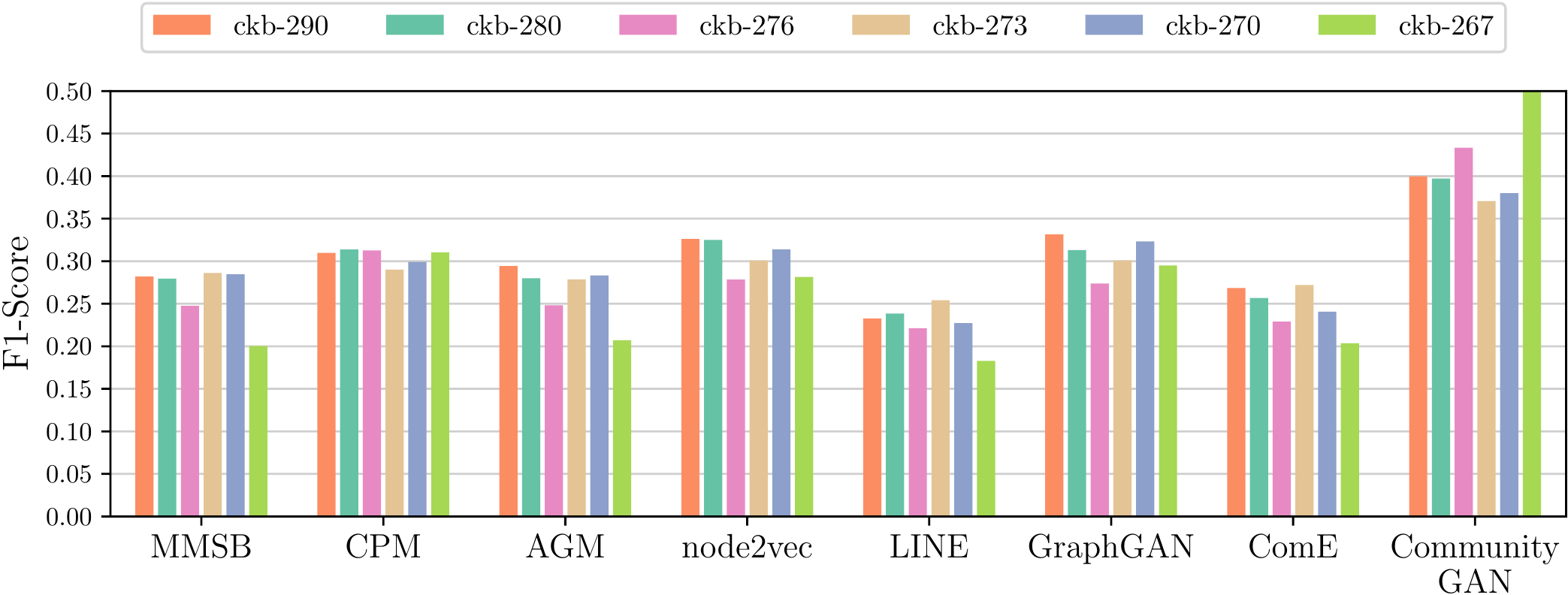}
\caption{Performance of methods on the synthetic graphs.}
\label{fig:synthetic-data-1}
\end{figure}

For each method and each dataset, we calculate the average value of F1-Score.
The performance of $8$ methods on the series of graphs is shown in Figure \ref{fig:synthetic-data-1}.
As we can see: 
1) \ComGAN~significantly outperforms all the baselines on all the synthetic graphs.
2) Although \ComGAN~is based on AGM and utilizes AGM as the pre-train method, with the minimax game competition between the discriminator and generator, \ComGAN~achieves huge improvements compared to AGM.
3) On the whole, the performance of $4$ graph representation learning methods (node2vec, LINE, GraphGAN, ComE) falls with the increase of overlapping density among communities, while $3$ traditional overlapping community detection methods (MMSB, CPM, AGM) keep relatively steady.
Among these $3$ methods, the only exception is that the performance of MMSB and AGM drops largely on ckb-267.
One possible reason is that the average degree of ckb-267 increases significantly compared to other graphs, and thus the performance of the two methods are affected.
However, although based on AGM, \ComGAN~still performs very well on the graph ckb-267 with the well designed positive/negative sampling and the minimax game between the generator and discriminator.

\subsection{Efficacy of Motif-Level Generation and Discrimination}

\begin{figure}[tbp]
\includegraphics[width=\linewidth]{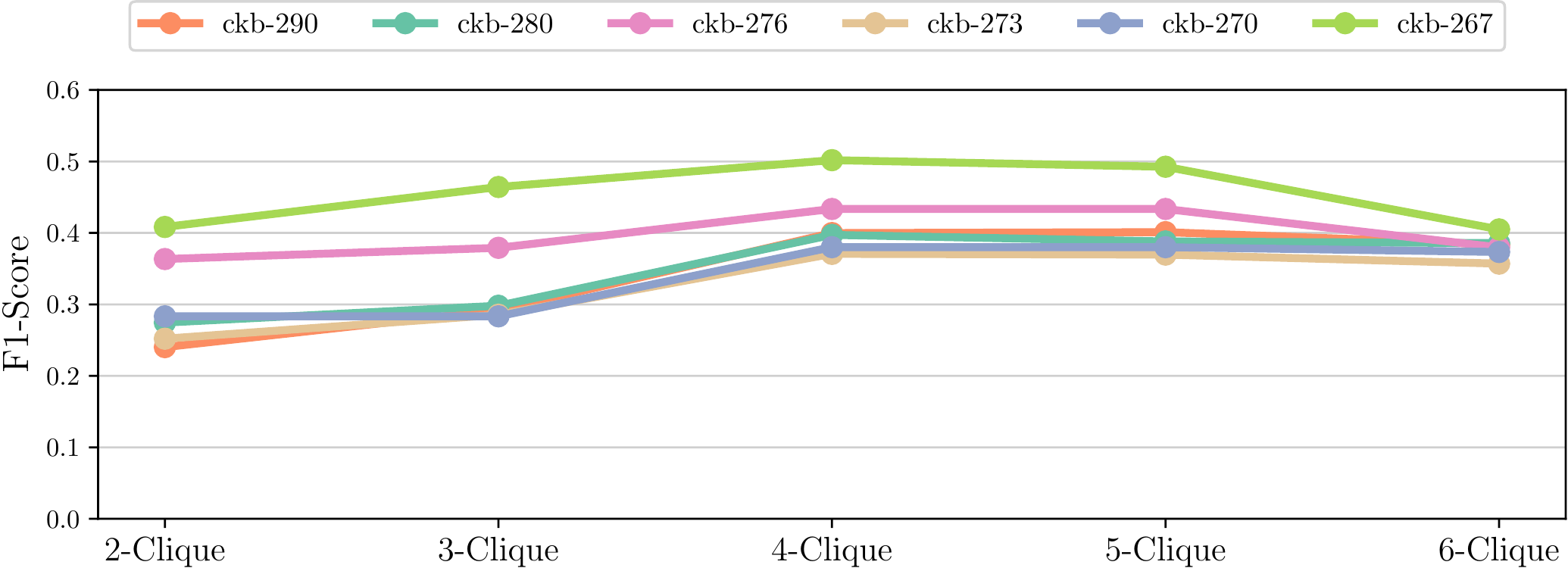}
\caption{Performance of \ComGAN~with different size of cliques on the synthetic graphs.}
\label{fig:synthetic-data-2}
\end{figure}

To evaluate the efficacy of the motif-level generation and discrimination, we tune the size of cliques in the training process of \ComGAN.
Figure \ref{fig:synthetic-data-2} shows the performance of \ComGAN~with different sizes of cliques on the series of synthetic graphs.
We can find that, generally, cliques with a size of $4$ or $5$ can reach the best performance.
It means that, compared to edges and triangles, the larger cliques in graphs correspond more to the community structure.
In other words, community detection methods should not only focus on the edges in the graph, but also pay more attention to the motifs, especially larger motifs.
However, we can see that in most graphs the performance falls when \ComGAN~generates and discriminates 6-vertex cliques.
Our explanation is that too-large cliques are rare in graphs, which can only cover a relatively small number of nodes.
Because of these uncovered nodes, few positive samples can be used in the training process, which in turn causes \ComGAN~to fail to learn appropriate representation vectors.

\begin{table}[tbp]
\centering
\caption{Statistics of cliques in the synthetic graphs.}
\label{tab:clique-count}
\begin{tabular}{cccccc}
\multicolumn{6}{c}{Number of cliques}                                                                                                                       \\ \hline
\multicolumn{1}{|l|}{Model}        & \multicolumn{1}{c|}{2-clique}       & \multicolumn{1}{c|}{3-clique}       & \multicolumn{1}{c|}{4-clique}       & \multicolumn{1}{c|}{5-clique}       & \multicolumn{1}{c|}{6-clique}       \\ \hline \hline
\multicolumn{1}{|l|}{ckb-290} & \multicolumn{1}{c|}{29491} & \multicolumn{1}{c|}{105585} & \multicolumn{1}{c|}{82617}  & \multicolumn{1}{c|}{31919}  & \multicolumn{1}{c|}{14354} \\ \hline
\multicolumn{1}{|l|}{ckb-280} & \multicolumn{1}{c|}{30055} & \multicolumn{1}{c|}{115047} & \multicolumn{1}{c|}{96661}  & \multicolumn{1}{c|}{33735}  & \multicolumn{1}{c|}{9539} \\ \hline
\multicolumn{1}{|l|}{ckb-276} & \multicolumn{1}{c|}{38775} & \multicolumn{1}{c|}{175639} & \multicolumn{1}{c|}{189300} & \multicolumn{1}{c|}{126720} & \multicolumn{1}{c|}{76247} \\ \hline
\multicolumn{1}{|l|}{ckb-273} & \multicolumn{1}{c|}{28025} & \multicolumn{1}{c|}{122566} & \multicolumn{1}{c|}{166760} & \multicolumn{1}{c|}{163765} & \multicolumn{1}{c|}{165575} \\ \hline
\multicolumn{1}{|l|}{ckb-270} & \multicolumn{1}{c|}{30740} & \multicolumn{1}{c|}{140547} & \multicolumn{1}{c|}{203114} & \multicolumn{1}{c|}{211604} & \multicolumn{1}{c|}{214018} \\ \hline
\multicolumn{1}{|l|}{ckb-267} & \multicolumn{1}{c|}{65813} & \multicolumn{1}{c|}{377686} & \multicolumn{1}{c|}{580950} & \multicolumn{1}{c|}{811522} & \multicolumn{1}{c|}{1097762} \\ \hline
\\
\multicolumn{6}{c}{Number of covered vertices}                                                                                                                       \\ \hline
\multicolumn{1}{|l|}{Model}        & \multicolumn{1}{c|}{2-clique}       & \multicolumn{1}{c|}{3-clique}       & \multicolumn{1}{c|}{4-clique}       & \multicolumn{1}{c|}{5-clique}       & \multicolumn{1}{c|}{6-clique}       \\ \hline \hline
\multicolumn{1}{|l|}{ckb-290} & \multicolumn{1}{c|}{1078} & \multicolumn{1}{c|}{1078} & \multicolumn{1}{c|}{1078} & \multicolumn{1}{c|}{1065} & \multicolumn{1}{c|}{595} \\ \hline
\multicolumn{1}{|l|}{ckb-280} & \multicolumn{1}{c|}{1044} & \multicolumn{1}{c|}{1044} & \multicolumn{1}{c|}{1044} & \multicolumn{1}{c|}{1034} & \multicolumn{1}{c|}{617} \\ \hline
\multicolumn{1}{|l|}{ckb-276} & \multicolumn{1}{c|}{1023} & \multicolumn{1}{c|}{1023} & \multicolumn{1}{c|}{1023} & \multicolumn{1}{c|}{1018} & \multicolumn{1}{c|}{714} \\ \hline
\multicolumn{1}{|l|}{ckb-273} & \multicolumn{1}{c|}{1018} & \multicolumn{1}{c|}{1018} & \multicolumn{1}{c|}{1018} & \multicolumn{1}{c|}{1012} & \multicolumn{1}{c|}{798} \\ \hline
\multicolumn{1}{|l|}{ckb-270} & \multicolumn{1}{c|}{1007} & \multicolumn{1}{c|}{1007} & \multicolumn{1}{c|}{1007} & \multicolumn{1}{c|}{1005} & \multicolumn{1}{c|}{797} \\ \hline
\multicolumn{1}{|l|}{ckb-267} & \multicolumn{1}{c|}{1123} & \multicolumn{1}{c|}{1123} & \multicolumn{1}{c|}{1123} & \multicolumn{1}{c|}{1120} & \multicolumn{1}{c|}{907} \\ \hline
\end{tabular}
\end{table}

To prove such an explanation, we have counted the number of cliques and number of vertices covered by such cliques, as shown in Table \ref{tab:clique-count}.
We can see that no matter whether the number of cliques increases or not, the number of covered vertices does not change very much from 2-clique to 5-clique.
Thus \ComGAN~can benefit from the larger cliques to gather better community structure, and then reaches the best performance when clique size is $4$ or $5$.
However, when the size of clique reaches 6, the number of covered vertices drops dramatically, and thus the performance of \ComGAN~decreases.

\subsection{Discussion}

With the two experiments on the series of synthetic graphs, we can draw the following conclusions:
\begin{itemize}
  \item With the same special vector design as AGM, \ComGAN~holds the ability to solve the dense overlapping problem. With the increase of overlapping among communities, the performance of \ComGAN~can maintain steadiness and does not fall like the baseline methods.
  \item With the minimax game competition between the discriminator and generator, \ComGAN~gains significant performance improvements based on AGM.
  \item With proper sizes, the motif-level generation and discrimination can help models learn better community structures compared to edge-level optimization.
\end{itemize}

\section{Real-world Scenarios}

To complement the previous experiments, in this section, we evaluate the performance of \ComGAN~on a series of real-world datasets.
Specifically, we choose two application scenarios for experiments, i.e., community detection and clique prediction.
\subsection{Datasets}

We evaluate our model using two categories of datasets, one with ground-truth communities and one without.

\vspace{5pt}\noindent\textbf{Datasets with ground-truth communities.\footnote{http://snap.stanford.edu/data/\#communities}}
Because of the training time of some baselines, we only sample three subgraphs with 100 ground-truth communities as the experiment networks from these three large networks.
\begin{description}
  \item[Amazon] is collected by crawling Amazon website.
  The vertices represent products; the edges indicate the frequently co-purchase relationships; the ground-truth communities are defined by the product categories in Amazon.
  This graph has 3,225 vertices and 10,262 edges.
  \item[Youtube] is a network of social relationships of Youtube website users.
  The vertices represent users; the edges indicate friendships among the users; the user-defined groups are considered as ground-truth communities.
  This graph has 4,890 vertices and 20,787 edges.
  \item[DBLP] is a co-authorship network from DBLP.
  The vertices represent researchers; the edges indicate the co-author relationships; authors who have published in a same journal or conference form a community.
  This graph has 10,824 vertices and 38,732 edges.
\end{description}

\begin{table}[tbp]
\centering
\caption{F1-Score and NMI on community detection.}
\label{tab:community_detection_result}
\begin{tabular}{cccc}
\multicolumn{4}{c}{F1-Score}
\\ \hline

\multicolumn{1}{|l|}{Model}          & \multicolumn{1}{c|}{Amazon}          & \multicolumn{1}{c|}{Youtube}         & \multicolumn{1}{c|}{DBLP}            \\ \hline \hline
\multicolumn{1}{|l|}{MMSB}           & \multicolumn{1}{c|}{0.366}           & \multicolumn{1}{c|}{0.124}           & \multicolumn{1}{c|}{0.104}           \\ \hline
\multicolumn{1}{|l|}{CPM}            & \multicolumn{1}{c|}{0.214}           & \multicolumn{1}{c|}{0.058}           & \multicolumn{1}{c|}{0.318}           \\ \hline
\multicolumn{1}{|l|}{AGM}            & \multicolumn{1}{c|}{0.711}           & \multicolumn{1}{c|}{0.158}           & \multicolumn{1}{c|}{0.398}           \\ \hline
\multicolumn{1}{|l|}{node2vec}       & \multicolumn{1}{c|}{0.550}           & \multicolumn{1}{c|}{0.288}           & \multicolumn{1}{c|}{0.265}           \\ \hline
\multicolumn{1}{|l|}{LINE}           & \multicolumn{1}{c|}{0.532}           & \multicolumn{1}{c|}{0.170}           & \multicolumn{1}{c|}{0.208}           \\ \hline
\multicolumn{1}{|l|}{GraphGAN}       & \multicolumn{1}{c|}{0.518}           & \multicolumn{1}{c|}{0.303}           & \multicolumn{1}{c|}{0.276}           \\ \hline
\multicolumn{1}{|l|}{ComE}           & \multicolumn{1}{c|}{0.562}           & \multicolumn{1}{c|}{0.213}           & \multicolumn{1}{c|}{0.240}           \\ \hline
\multicolumn{1}{|l|}{\ComGAN}        & \multicolumn{1}{c|}{\textbf{0.860}}  & \multicolumn{1}{c|}{\textbf{0.327}}  & \multicolumn{1}{c|}{\textbf{0.456}}  \\ \hline
\\
\multicolumn{4}{c}{NMI}
\\ \hline
\multicolumn{1}{|l|}{Model}          & \multicolumn{1}{c|}{Amazon}          & \multicolumn{1}{c|}{Youtube}         & \multicolumn{1}{c|}{DBLP}           \\ \hline \hline
\multicolumn{1}{|l|}{MMSB}            & \multicolumn{1}{c|}{0.068}           & \multicolumn{1}{c|}{0.031}           & \multicolumn{1}{c|}{0.000}           \\ \hline
\multicolumn{1}{|l|}{CPM}            & \multicolumn{1}{c|}{0.027}           & \multicolumn{1}{c|}{0.000}           & \multicolumn{1}{c|}{0.066}           \\ \hline
\multicolumn{1}{|l|}{AGM}           & \multicolumn{1}{c|}{0.635}           & \multicolumn{1}{c|}{0.025}           & \multicolumn{1}{c|}{0.059}           \\ \hline
\multicolumn{1}{|l|}{node2vec}       & \multicolumn{1}{c|}{0.370}           & \multicolumn{1}{c|}{0.071}           & \multicolumn{1}{c|}{0.068}           \\ \hline
\multicolumn{1}{|l|}{LINE}           & \multicolumn{1}{c|}{0.248}           & \multicolumn{1}{c|}{0.070}           & \multicolumn{1}{c|}{0.027}           \\ \hline
\multicolumn{1}{|l|}{GraphGAN}       & \multicolumn{1}{c|}{0.417}           & \multicolumn{1}{c|}{0.049}           & \multicolumn{1}{c|}{0.083}           \\ \hline
\multicolumn{1}{|l|}{ComE}           & \multicolumn{1}{c|}{0.413}           & \multicolumn{1}{c|}{\textbf{0.091}}  & \multicolumn{1}{c|}{0.059}           \\ \hline
\multicolumn{1}{|l|}{\ComGAN}        & \multicolumn{1}{c|}{\textbf{0.853}}  & \multicolumn{1}{c|}{\textbf{0.091}}  & \multicolumn{1}{c|}{\textbf{0.153}}  \\ \hline
\end{tabular}
\end{table}

\vspace{5pt}\noindent\textbf{Datasets without ground-truth communities.\footnote{http://snap.stanford.edu/data/\#canets}}
\begin{description}
  \item[arXiv-AstroPh] is from the e-print arXiv and covers scientific collaborations between authors with papers submitted to the Astro Physics category.
  The vertices represent authors, and the edges indicate co-author relationships.
  This graph has 18,772 vertices and 198,110 edges.
  \item[arXiv-GrQc] is also from arXiv and covers scientific collaborations between authors with papers submitted to the General Relativity and Quantum Cosmology categories.
  This graph has 5,242 vertices and 14,496 edges.
\end{description}

\subsection{Community Detection}
The community detection experiment is conducted on three networks with ground-truth community memberships, which allows us to quantify the accuracy of community detection methods by evaluating the level of correspondence between detected and ground-truth communities.

\vspace{5pt}\noindent\textbf{Setup.}
In the community detection experiment, all the baselines and setups are the same as the synthetic data experiments, whose details refer to \S \ref{subsec:experiment-setup}.

\vspace{5pt}\noindent\textbf{Evaluation Metric.}
Besides the F1-Score described in synthetic dataset experiments (refer to \S \ref{subsection:evaluation metric}), to comprehensively evaluate the performance of methods, we also introduce the Normalized Mutual Information (NMI) in the community detection experiment.
NMI is a measure of similarity borrowed from information theory.
Later, it is extended to measure the quality of overlapping communities.
Similar to F1-Score, the higher NMI means the better performance.
Please refer to \cite{NMI} for details.

\vspace{5pt}\noindent\textbf{Results.}
Table \ref{tab:community_detection_result} shows the F1-Score and NMI of 8 methods on 3 networks, respectively.
As we can see:
1) Performance of CPM, MMSB and LINE is relatively poor in community detection, which means they cannot capture the pattern of community structure in graphs well.
2) Node2vec, ComE and GraphGAN perform better than CPM, MMSB and LINE.
One possible reason is that node2vec and GraphGAN adopt the random walk to capture the vertex context which is similar to community structure in some way, and ComE solves the network embedding and community detection task in one unified framework.
3) \ComGAN~outperforms all the baselines (including its pre-train method AGM) in community detection task.
Specifically, \ComGAN~improves F1-Score by 7.9\% (on Youtube) to 21.0\% (on Amazon) and improves NMI by at most 34.3\% (on Amazon).
Our explanation is that the consideration of dense community overlapping provides \ComGAN~a higher learning flexibility than the non-overlapping or sparse overlapping community detection baselines.
Moreover, the minimax game between the generator and discriminator drives \ComGAN~to gain significant performance improvements based on AGM.

\begin{figure}[t]
\centering
\subfigure[Generator]{
  \begin{minipage}{0.47\linewidth}
  \centering
  \includegraphics[width=\linewidth]{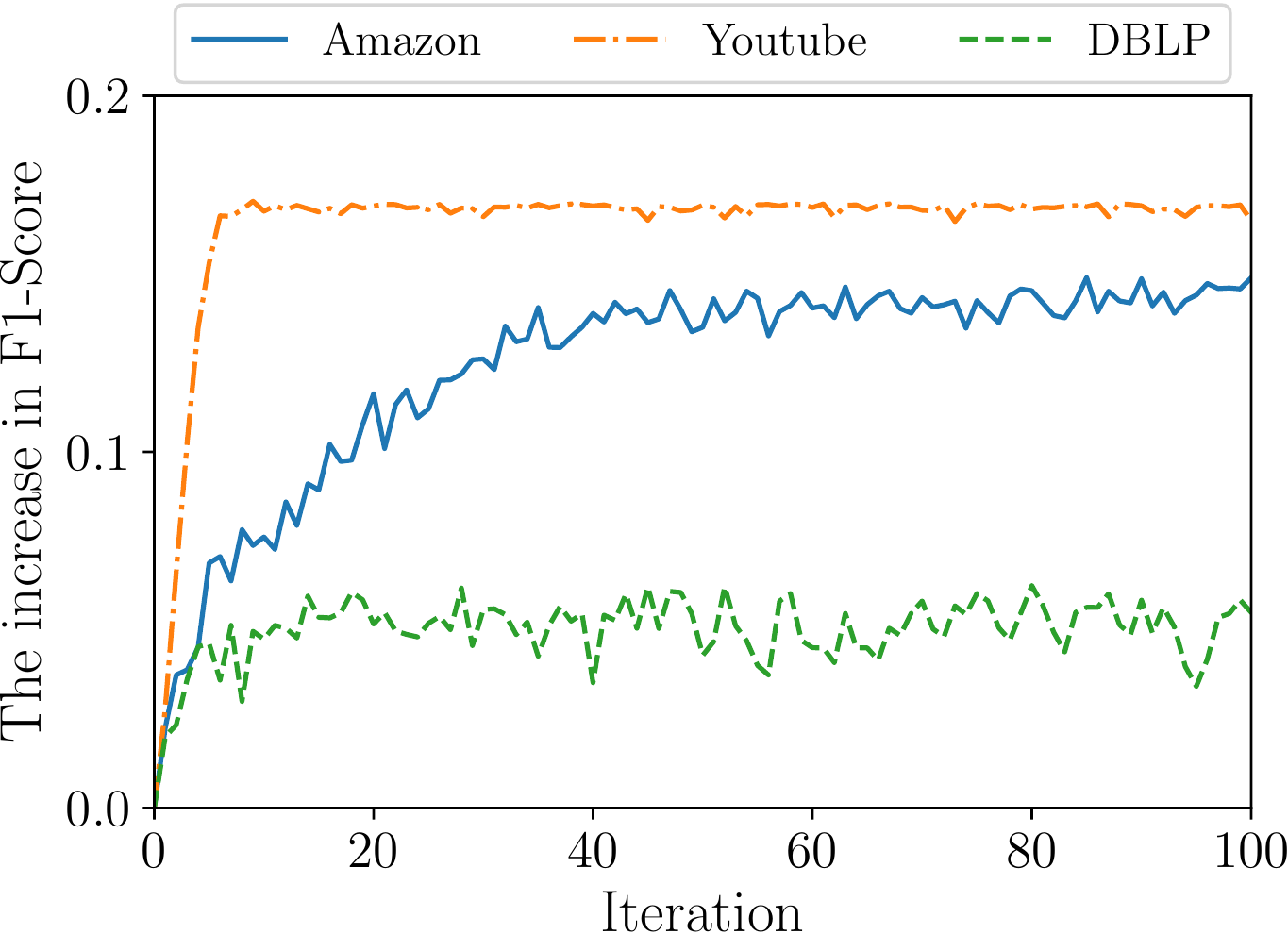}
  \end{minipage}
}
\subfigure[Discriminator]{
  \begin{minipage}{0.47\linewidth}
  \centering
  \includegraphics[width=\linewidth]{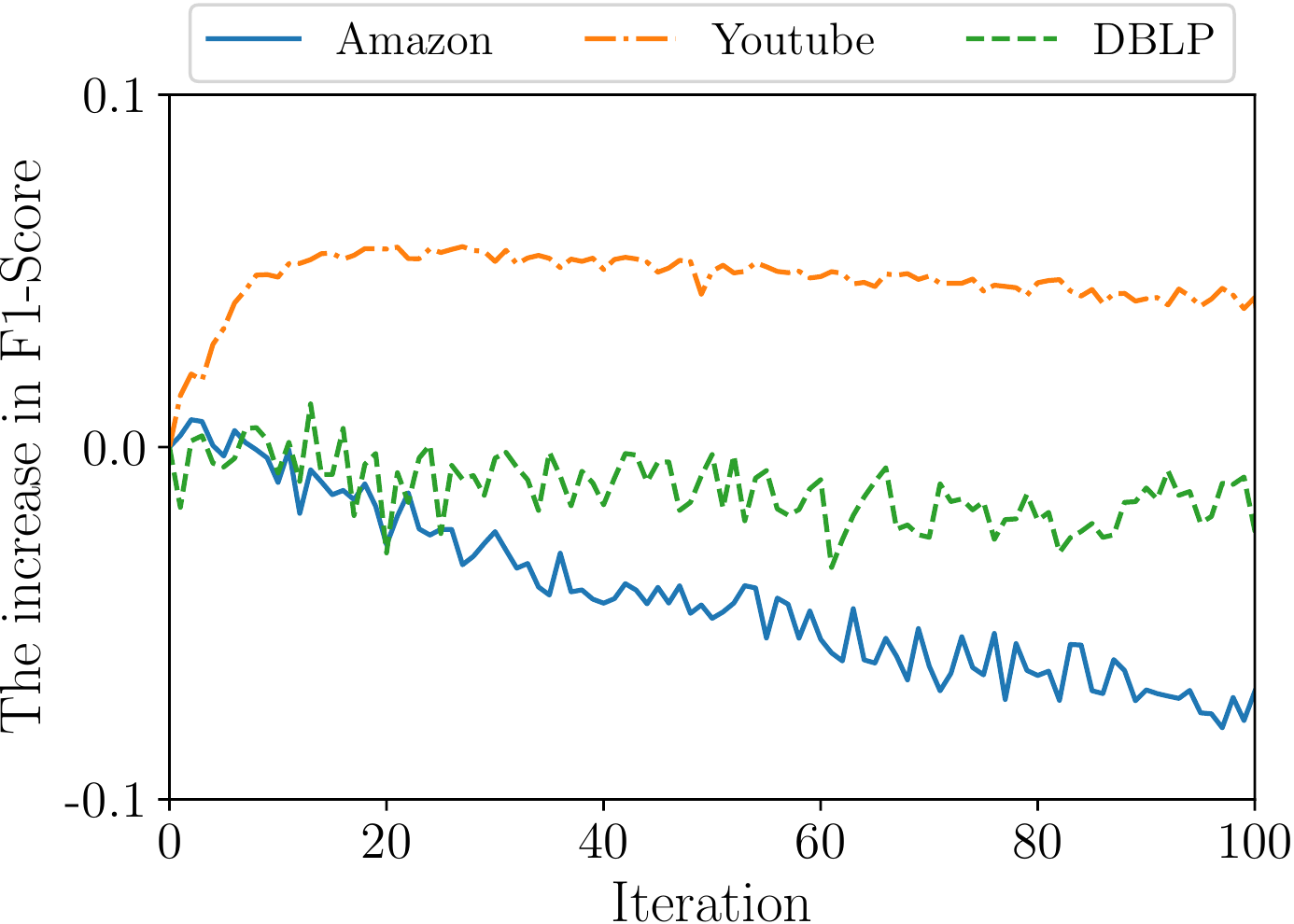}
  \end{minipage}
}
\caption{Learning curves of the generator and the discriminator of \ComGAN~in community detection.}
\label{fig:learning-curve}
\end{figure}

To intuitively understand the learning stability of \ComGAN, we further illustrate the learning curves of the generator and the discriminator on the three datasets in Figure \ref{fig:learning-curve}.
Because the F1-Score on the three datasets differs very much, we only demonstrate the relative increase in the score.
As we can see, the generator performs outstandingly well after convergence, while the performance of the discriminator falls a little or boosts at first and then falls.
Maybe it is because that the discriminator aims to distinguish ground truth from generated samples better.
However, the generated samples are all around the center vertex, which causes the discriminator to lose some global discrimination ability.

\begin{table}[tbp]
\centering
\caption{AUC on clique prediction.}
\label{tab:clique_prediction_result}
\begin{tabular}{cccc}
\multicolumn{4}{c}{arXiv-AstroPh}                                                                                                                       \\ \hline
\multicolumn{1}{|l|}{Model}        & \multicolumn{1}{c|}{2-clique}       & \multicolumn{1}{c|}{3-clique}       & \multicolumn{1}{c|}{4-clique}       \\ \hline \hline
\multicolumn{1}{|l|}{AGM}     & \multicolumn{1}{c|}{0.919}          & \multicolumn{1}{c|}{0.987}          & \multicolumn{1}{c|}{0.959}          \\ \hline
\multicolumn{1}{|l|}{node2vec}     & \multicolumn{1}{c|}{0.579}          & \multicolumn{1}{c|}{0.514}          & \multicolumn{1}{c|}{0.544}          \\ \hline
\multicolumn{1}{|l|}{LINE}         & \multicolumn{1}{c|}{0.918}          & \multicolumn{1}{c|}{0.980}          & \multicolumn{1}{c|}{0.963}          \\ \hline
\multicolumn{1}{|l|}{GraphGAN}     & \multicolumn{1}{c|}{0.799}          & \multicolumn{1}{c|}{0.859}          & \multicolumn{1}{c|}{0.855}          \\ \hline
\multicolumn{1}{|l|}{ComE}         & \multicolumn{1}{c|}{0.904}          & \multicolumn{1}{c|}{0.951}          & \multicolumn{1}{c|}{0.953}          \\ \hline
\multicolumn{1}{|l|}{\ComGAN}      & \multicolumn{1}{c|}{\textbf{0.923}}          & \multicolumn{1}{c|}{\textbf{0.990}} & \multicolumn{1}{c|}{\textbf{0.970}} \\ \hline
\\
\multicolumn{4}{c}{arXiv-GrQc}                                                                                                                       \\ \hline
\multicolumn{1}{|l|}{Model}        & \multicolumn{1}{c|}{2-clique}       & \multicolumn{1}{c|}{3-clique}       & \multicolumn{1}{c|}{4-clique}       \\ \hline \hline
\multicolumn{1}{|l|}{AGM}     & \multicolumn{1}{c|}{0.900}          & \multicolumn{1}{c|}{0.980}          & \multicolumn{1}{c|}{0.871}          \\ \hline
\multicolumn{1}{|l|}{node2vec}     & \multicolumn{1}{c|}{0.632}          & \multicolumn{1}{c|}{0.569}          & \multicolumn{1}{c|}{0.534}          \\ \hline
\multicolumn{1}{|l|}{LINE}         & \multicolumn{1}{c|}{\textbf{0.969}} & \multicolumn{1}{c|}{0.989}          & \multicolumn{1}{c|}{0.880}          \\ \hline
\multicolumn{1}{|l|}{GraphGAN}     & \multicolumn{1}{c|}{0.756}          & \multicolumn{1}{c|}{0.880}          & \multicolumn{1}{c|}{0.728}          \\ \hline
\multicolumn{1}{|l|}{ComE}         & \multicolumn{1}{c|}{0.924}          & \multicolumn{1}{c|}{0.962}          & \multicolumn{1}{c|}{0.914}          \\ \hline
\multicolumn{1}{|l|}{\ComGAN}      & \multicolumn{1}{c|}{0.904}          & \multicolumn{1}{c|}{\textbf{0.993}} & \multicolumn{1}{c|}{\textbf{0.956}} \\ \hline
\end{tabular}
\end{table}

\subsection{Clique Prediction}
In clique prediction, our goal is to predict whether a given subset of vertices is a clique.
Therefore, this task shows the performance of graph local structure extraction ability of different graph representation learning methods.

\vspace{5pt}\noindent\textbf{Setup.}
In the clique prediction experiment, because some traditional community detection methods (including MMSB and CPM) cannot predict the existence of edges among vertices, these methods are omitted in this experiment.
To analyze the effect of motif generation and discrimination in \ComGAN, in this experiment, we have evaluated the prediction for 2-clique (same to edge), 3-clique and 4-clique.
With the size of cliques determined, we randomly hide some cliques, which cover 10\% of edges, in the original graph as ground truth, and use the left graph to train all graph representation learning models.
After training, we obtain the representation vectors for all vertices and use logistic regression method to predict the probability of being clique for a given vertex set.
Our test set consists of the hidden vertex sets (cliques) in the original graph as the positive samples and the randomly selected non-fully connected vertex sets as negative samples with the equal number.

\vspace{5pt}\noindent\textbf{Results.}
We use arXiv-AstroPh and arXiv-GrQc as datasets, and report the results of AUC in Table \ref{tab:clique_prediction_result}.
As we can see, even though in 2-clique (edge) prediction \ComGAN~does not always get the highest score, \ComGAN~outperforms all the baselines in both 3-clique and 4-clique prediction.
For example, on arXiv-GrQc, \ComGAN~achieves gains of 0.40\% to 74.52\% and 4.60\% to 79.03\% in 3-clique and 4-clique prediction, respectively.
This indicates that though \ComGAN~is designed for community detection, with the design of motif generation and discrimination, it can still effectively encode the information of clique structures into the learned representations.

\subsection{Efficiency analysis}
\label{efficiency-analysis}
\begin{figure}[tbp]
\includegraphics[width=\linewidth]{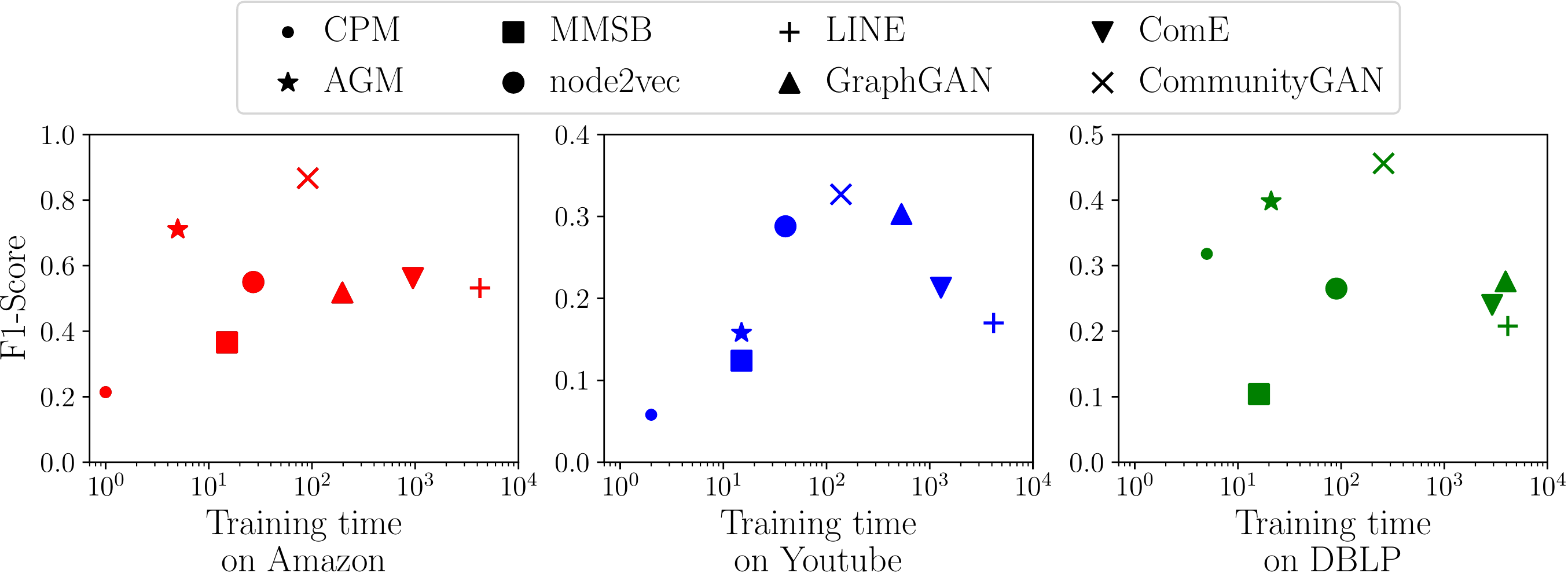}
\caption{Performance as training time (Sec.).}
\label{fig:efficiency}
\end{figure}

In this paper, we propose Graph AGM for efficiently generating the most likely motifs with graph structure awareness.
Because of the random walk process, in which the exact time complexity is not easy to infer, we evaluate the efficiency of \ComGAN~by directly comparing the training time with baselines.
In this evaluation, the number of threads is set as 16 if the model supports parallelization and other parameters for baselines are as default.
Figure \ref{fig:efficiency} illustrates the performance and training time.
Notably, the training time of \ComGAN~includes the time of the pre-training process.
Even though \ComGAN~is not the fastest model, the training time of \ComGAN~is still acceptable and its performance significantly outperforms the faster models.

\section{Conclusion}
In this paper we proposed \ComGAN~that jointly solves the overlapping community detection and graph representation learning.
Unlike the embedding of general graph learning algorithms in which the vector values have no meanings, the embedding in \ComGAN~indicates the membership strength of vertices to communities, which enables \ComGAN~to detect densely overlapped communities.
Then a specifically designed GAN is adopted to optimize such embedding.
Through the minimax game of motif-level generator and discriminator, both of them can boost their performance and finally output better community structures.
We adopted CKB Graph Generator to create a series of synthetic graphs with ground-truth communities.
Two experiments were conducted on these graphs to prove the ability of \ComGAN~to solve dense overlapping problem and its efficacy of motif generation and discrimination.
Additionally, to complement the experiments on the synthetic datasets, we did experiments on five real-world datasets in two scenarios, where the results demonstrate that \ComGAN~substantially outperforms baselines in all experiments due to its specific embedding design and motif-level optimization.

\section{Acknowledgements}
The work is supported by National Key R\&D Program of China (2018YFB1004702), National Natural Science Foundation of China (61532012, 61829201, 61702327, 61772333), Shanghai Sailing Program (17YF1428200).
Weinan Zhang and Xinbing Wang are the corresponding authors.
\bibliographystyle{ACM-Reference-Format}
\balance 
\bibliography{main}

\end{document}